\def\year{2022}\relax
\documentclass[letterpaper]{article} 
\usepackage{aaai22}  
\usepackage{times}  
\usepackage{helvet}  
\usepackage{courier}  
\usepackage[hyphens]{url}  
\usepackage{graphicx} 
\usepackage{natbib}  
\usepackage{caption} 
\DeclareCaptionStyle{ruled}{labelfont=normalfont,labelsep=colon,strut=off} 
\usepackage{algorithm}
\usepackage{algorithmic}

\usepackage{newfloat}
\usepackage{listings}
\floatstyle{ruled}
\newfloat{listing}{tb}{lst}{}
\floatname{listing}{Listing}
\usepackage[dvipsnames]{xcolor}
\usepackage{xspace}
\usepackage{threeparttable}
\usepackage{multirow}
\usepackage{graphicx}
\usepackage{tabularx}
\usepackage{mathtools}
\usepackage{amsmath}
\usepackage{booktabs}
\usepackage{multicol}
\usepackage{multirow}
\usepackage{dsfont}
\usepackage{subcaption}
\newcommand{\figref}[2][]{Figure#1~\ref{#2}\xspace}
\newcommand{\tabref}[2][]{Table#1~\ref{#2}\xspace}

\renewcommand{\vec}[1]{\boldsymbol{#1}}

\newcommand{\class}[1]{\textsc{#1}\xspace}
\newcommand{\aae}{\class{AAE}\xspace}
\newcommand{\sae}{\class{SAE}\xspace}
\newcommand{\happy}{\class{happy}\xspace}
\newcommand{\sad}{\class{sad}\xspace}
\newcommand{\nurse}{\class{nurse}\xspace}
\newcommand{\surgeon}{\class{surgeon}\xspace}

\newcommand{\dataset}[1]{\textbf{#1}\xspace}
\newcommand{\hatespeech}{\dataset{Hate speech}}
\newcommand{\imsitu}{\dataset{imSitu}}
\newcommand{\moji}{\dataset{Moji}}
\newcommand{\bios}{\dataset{Bios}}

\newcommand{\bert}{BERT\xspace}
\newcommand{\deepmoji}{DeepMoji\xspace}

\newcommand{\method}[1]{\textsf{#1}\xspace}
\newcommand{\vanilla}{\method{CE}}
\newcommand{\diffadv}{\method{Adv}}
\newcommand{\inlp}{\method{INLP}}
\newcommand{\contrastive}{\method{Con$_{*}$}}
\newcommand{\ceadd}{\method{Con$_{\text{ce}+\text{scl}}$}}
\newcommand{\ceminus}{\method{Con$_{\text{ce}-\text{fcl}}$}}

\newcommand{\tuneboth}{\method{Con$^{\mathrm{ft}}_{*}$}}

\newcommand{\clattr}{\ensuremath{\mathcal{L}_{\text{fcl}}}\xspace}
\newcommand{\clmain}{\ensuremath{\mathcal{L}_{\text{scl}}}\xspace}
\newcommand{\clce}{\ensuremath{\mathcal{L}_{\text{ce}}}\xspace}

\newcommand{\pareto}{Pareto\xspace}
\newcommand{\accuracy}{Accuracy\xspace}
\newcommand{\hidden}{Leakage@\textbf{h}\xspace}
\newcommand{\logits}{Leakage@$\hat{\textbf{y}}$\xspace}
\newcommand{\gap}{GAP\xspace}

\newcommand{\rank}{Tradeoff\xspace}
\newcommand{\gputime}{Time\xspace}

\newcommand{\backbone}[1]{\textit{\texttt{Embed({#1})}}\xspace}
\newcommand{\encoder}[1]{\textit{\texttt{Enc({#1})}}\xspace}

\title{Contrastive Learning for Fair Representations}
\title{Contrastive Learning for Fair Representations}
\author {
	Aili Shen,
	Xudong Han,
	Trevor Cohn,
	Timothy Baldwin,
	Lea Frermann
}
\affiliations {
	School of Computing and Information Systems\\
	The University of Melbourne \\
	Victoria 3010, Australia\\
	\{aili.shen, t.cohn, tbaldwin, lfrermann\}@unimelb.edu.au, xudongh1@student.unimelb.edu.au
}

\begin{document}
	
	\maketitle
	
	\begin{abstract}
		
		Trained classification models can unintentionally lead to biased representations and predictions, which can reinforce societal preconceptions and stereotypes. 
		Existing debiasing methods for classification models, such as adversarial training, are often expensive to train and difficult to optimise. 
		In this paper, we propose a method for mitigating bias in classifier training by incorporating contrastive learning, in which instances sharing the same class label are encouraged to have similar representations, while instances sharing a protected attribute are forced further apart.
		In such a way our method learns representations which capture the task label in focused regions, while ensuring the protected attribute has diverse spread, and thus has limited impact on prediction and thereby results in fairer models. 
		Extensive experimental results across four tasks in NLP and computer vision show (a) that our proposed method can achieve fairer representations and realises bias reductions compared with competitive baselines; and (b) that it can do so without sacrificing main task performance; (c) that it sets a new state-of-the-art performance in one task despite reducing the bias. 
		Finally, our method is conceptually simple and agnostic to network architectures, and incurs minimal additional compute cost.

		

	\end{abstract}
	
	\section{Introduction}
	\label{sec:intro}
	
	Neural methods have achieved great success for classification tasks in NLP and computer vision. 
	However, datasets which neural models are trained on embody cultural and societal stereotypes from the real world. 
	Models trained on such datasets often capture spurious correlations between target labels and protected attributes, leading to biased predictions (i.e., models perform unequally for different sub-groups) and leakage of authorship-related sensitive information from learned representations (i.e., attackers can recover the demographic information from learned representations). 
	This kind of unfairness has been identified in various tasks, such as twitter sentiment analysis \cite{Blodgett:16,Han:21}, part-of-speech tagging \cite{Hovy:15,Li:18,Han:21b}, and image activity recognition \cite{Wang:19,Zhao:17}.

	To mitigate bias associated with protected attributes, various kinds of methods have been proposed \cite{Zhao:18,Zhao:17,Li:18}. 
	Data manipulation, such as balancing the dataset with respect to the protected attribute \cite{Wang:19} and augmenting a gender-biased dataset with gender-swapped sentences \cite{Zhao:18}, can reduce bias 
	at the input level, however it can be costly in terms of time and compute resources. 
	And it has been demonstrated that it is not an effective way to reduce bias. 
	Adversarial training is a popular method for mitigating bias by preventing a discriminator from reverse engineering protected attribute information from learned representations \cite{Elazar:18,Resheff:19,Han:21,Han:21b,Li:18}. 
	However, it is often difficult to optimise and increases model complexity and, consequently, computational cost.
	
	We propose a novel debiasing method based on contrastive learning \cite{Oord:18,Li:20,Tian:20,Henaff:20,Bui:21,Li:21,Chen:20b}. 
	Driven by the intuition that good and fair representations for classification should pull instances together {\it only} if they belong to the same class but not based on shared protected attributes (such as gender or race),
	we present an effective debiasing method based on  contrastive learning. 
	Specifically, our proposed method combines two contrastive loss components with a  cross-entropy loss, thereby maximising the similarities of instance pairs which share a main task label and minimising the similarities of such pairs from the protected attribute perspective.
	To the best of our knowledge, our work is the first to integrate  contrastive loss components 
	to obtain fairer representations. We demonstrate the effectiveness of our method  across four tasks, spanning NLP and computer vision. 
	Our contributions in this work are:
	\begin{itemize}
		\item We present a debiasing method based on  contrastive learning, combining cross-entropy loss with two  contrastive loss components; 
		\item Experimental results over four NLP and computer vision tasks show that our proposed method achieves the best accuracy--fairness tradeoff in each case; 
		\item Our method is simple to implement and agnostic to model architectures, and incurs minimal additional computing cost. 
	\end{itemize}
	
	\section{Related Work}
	\label{sec:related}
	We briefly review research in the two most related areas: debiasing methods and contrastive learning.

	\subsection{Debiasing Methods}
	Prior debiasing methods fall into three categories. First, data manipulation aims to balance the input, followed by re-training the model on a fairer dataset~\cite{Wang:19,Badjatiya:19,De-Arteaga:19,Elazar:18}. However, it has been shown to be both computationally prohibitive for large datasets or models, and ineffective in ensuring fair models~\cite{De-Arteaga:19,Wang:19}. Second,  post-processing methods ``bleach'' sensitive information from learnt representations after main task training. For example, Iterative Null Space Projection (INLP; \citet{Shauli:20}) iteratively trains a linear discriminator over a protected attribute given pre-computed fixed representations. The representations are then projected onto the linear discriminator's null-space, thereby making it difficult for a linear classifier to identify the protected attribute. In the third category, approaches augment the original training objective, to encourage the model to learn representations that are oblivious to protected attributes. Adversarial models are the prime example~\cite{Li:18,Zhang:18,Resheff:19,Wang:19,Barrett:19,Han:21}, in leveraging one~\citep{Li:18,Elazar:18} or more~\citep{Han:21} discriminators to encourage the main model to learn representations that do not reveal  protected information. Our method also introduces an augmented objective, however, unlike adversarial methods, it does not add additional model parameters, and hence is computationally much lighter weight. We compare our method against INLP and adversarial baselines.

	\subsection{Contrastive Learning}
	The basic idea behind contrastive learning (CL) is to pull similar instances together and push dissimilar instances apart by maximising the similarities of similar instances and minimising those of dissimilar pairs within the unit feature space~\cite{Oord:18,Tian:20,Li:20,Grill:20,Chen:20,Henaff:20}. CL has been particularly successful in computer vision, where positive (similar) instance pairs can be generated via data augmentation (i.e., systematic, meaning-invariant manipulation of an input image such as cropping or blurring~\cite{Chen:20,Fang:20,Cubuk:19}), and negative (dissimilar) instance pairs correspond to different items in the original data. More recently, supervised contrastive learning (SCL) was proposed in the context of classification, where positive instances belong to the same class, and negative instances belong to different classes~\cite{Khosla:20}. SCL, when combined with a cross entropy loss, has been shown to improve model robustness to noise and data sparsity~\cite{Beliz:21} as well as adversarial attacks~\citep{Bui:21}. We adapt SCL to {\it fair} supervised learning, and present evidence of its effectiveness in learning debiased representations and fair classifiers. 

	\section{Fair \& Supervised Contrastive Learning}
	\label{sec:method}
	
\begin{figure}[!t]
		\scalebox{1.0}{
			\includegraphics[width=\linewidth]{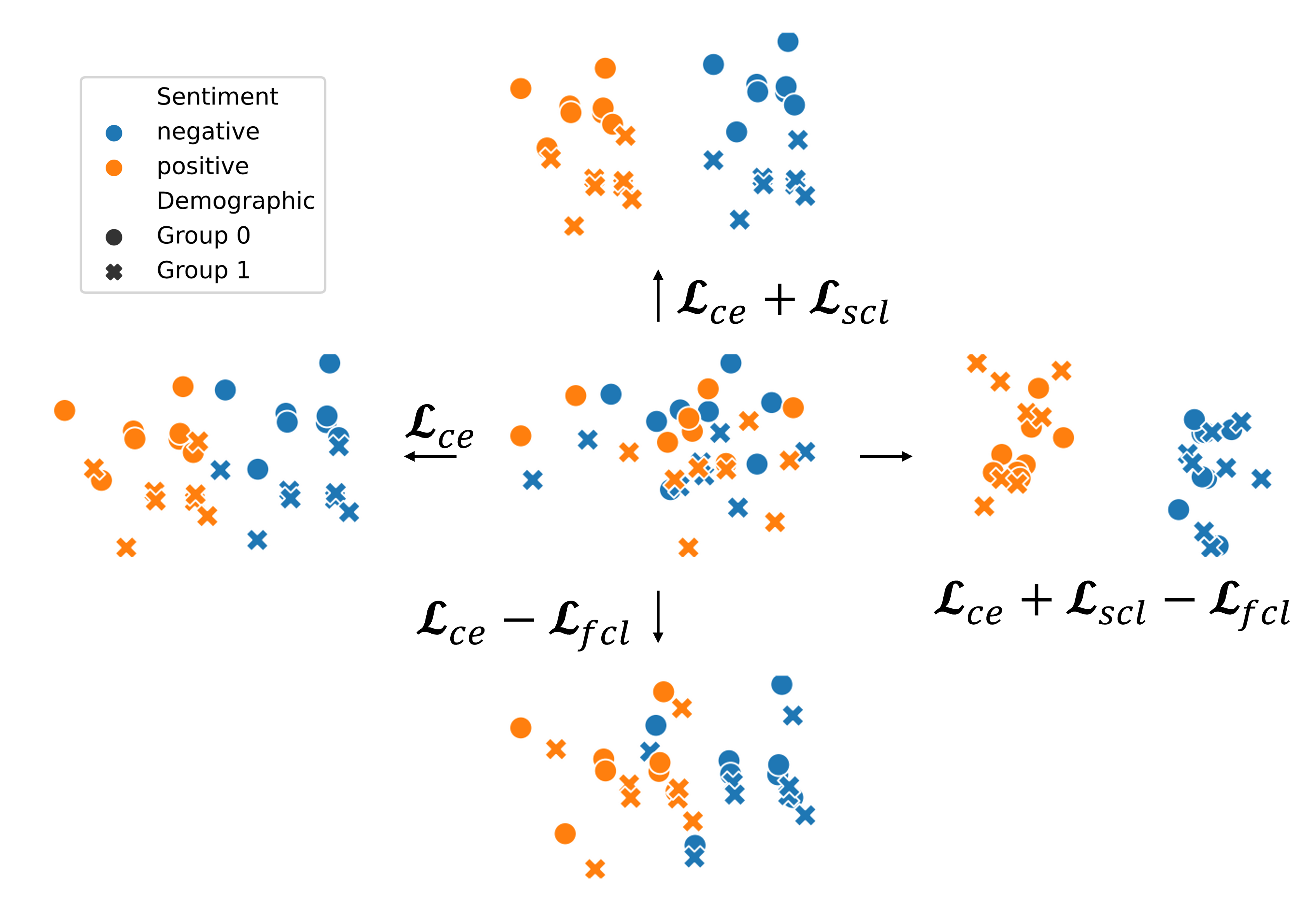}
		}
		\caption{Illustration of our proposed method in the context of sentiment classification, where \clce is cross-entropy loss, \clmain is contrastive loss based on main task, and \clattr is contrastive loss based on the protected attribute. } 		
		\label{illustration_method}
	\end{figure}

	Our proposed method equips supervised contrastive learning with an improved loss function which simultaneously encourages data separation in terms of the main class labels, and discourages the differentiation of data points on the basis of their protected attributes. Fair contrastive learning is illustrated in \figref{illustration_method}, and is compatible with different classifier architectures and data modalities, such as language and vision. Our architecture consists of three components: 
	\begin{enumerate}
		\item An \emph{embedding module}, $\vec{e}=$\backbone{$\vec{x}$}, which maps an input instance $\vec{x}$ (e.g., a document or an image) to a vector representation $\vec{e}$, which is in turn used as input to the encoder network;
		\item An \emph{encoder network}, $\vec{h}=$\encoder{$\vec{e}$}, which maps the input representation to the final hidden representation; 
		\item An \emph{aggregated objective} ($\mathcal{L}_{*}$), which is a weighted combination of a cross-entropy loss, contrastive loss based on main task labels, and contrastive loss based on protected attribute labels. 
	\end{enumerate} 
	
	
	\subsection{Cross-entropy Loss}

	The cross-entropy loss is defined as
	\begin{equation}
	\clce =-\dfrac{1}{N}\sum_{i=1}^{N}\sum_{c=1}^{Y}y_{i,c}\log{\hat{y}_{i,c}},
	\label{ce_loss}
	\end{equation}
	where $Y$ is the number of main task classes; $y_{i,c}$ denotes that the $i$th instance belongs to the main task class $c$; $\hat{y}_{i,c}$ denotes the predicted probability of the $i$th instance belonging to the main task class $c$; and
	$\hat{y}_{i,c}$ is the {softmax} output from the classifier, whose input is $\vec{h}$. 
	However, cross-entropy loss focuses only on maximising the predicted probability of the $i$th instance belonging to the gold-standard class, and not on how similar versus dissimilar instances are located in representation space. 
	In this work, we explicitly model the similarity of instances in the representation space via supervised contrastive learning.

	\subsection{Contrastive Losses}
	
	Given a mini-batch with a set of $N$ randomly sampled instances, 
	positive instance pairs (those which are truly equivalent) and negative instance pairs (those representing distinct concepts) are formed. We use two different criteria for creating these pairs: their main task label, and their protected attribute, as described below. Assuming a batch of positive and negative pairs, the contrastive loss is computed as,
	\begin{equation}
	\! \mathcal{L}_{\text{scl}} = \sum_{i=1}^{N}\dfrac{-1}{|P(i)|}\sum_{p\in P(i)}\mathrm{\log}\dfrac{\exp(\vec{\tilde{h}}_{i}\cdot\vec{\tilde{h}}_{p}/\tau)}{\sum_{q\in Q(i)}\exp(\vec{\tilde{h}}_{i}\cdot\vec{\tilde{h}}_{q}/\tau)} \, ,
	\label{contrast_definition}
	\end{equation}
	where $i{=}1\ldots N$ is the index of an instance in the mini-batch, 
	and $Q(i)\equiv\{1\ldots N\}\setminus\{i\}$; 
	\mbox{$\vec{\tilde{h}}_{i}=l_{2}(\encoder{\backbone{$\vec{x}_{i}$}})$} is the normalised representation; and $\tau>0$ is a scalar temperature parameter controlling softness. 
	$P(i)\equiv\{p\in Q(i): y^{p}=y^{i}\}$ is the set of instances that result in positive pairs with the $i$th instance, and $|P(i)|$ is its cardinality. We next describe how positive/negative pairs are created.
	
	\paragraph{Supervised Contrastive Loss:} \clmain is computed on positive and negative samples constructed based on main task labels (e.g., \class{pos} vs \class{neg} sentiment), where instances in the mini-batch belonging to the same main task class are used to construct positive samples; otherwise, they are used to form negative samples. 
	The intuition behind this loss component is that representations that are well-separated for the main task are more desirable, as illustrated in the top quadrant of \figref{illustration_method}, where the main task labels are indicated in blue and orange, and are separated into distinct clusters.

	\paragraph{Fair Contrastive Loss:} \clattr is based on positive and negative samples with respect to protected attribute labels (e.g., \class{male} vs \class{female}), where instances belonging to the same protected attribute class form positive samples; otherwise, they are used to construct negative samples. Our goal is to infer latent representations which are oblivious to the protected attribute of an instance.
	We enforce representations of instances with different protected attribute values to mix together by discouraging the model from effectively contrasting those instances, with the goal of reducing the correlation between the main task and protected attribute. This intuition is illustrated in the bottom quadrant of \figref{illustration_method}.

	\subsection{Objective Function}
	Our final objective 
	incorporates both contrastive learning methods, to produce task-indicative and protected-attribute-agnostic representations, as illustrated in the right quadrant of \figref{illustration_method}. It is formulated as a weighted average of \clce, \clmain, and \clattr, 
	\begin{equation}
	\mathcal{L}_{*}=\alpha\clce+\beta\{\text{\clmain}-\text{\clattr}\} \, .
	\label{overall_loss}
	\end{equation}
	The second term, \clmain, pulls instances from the same main task label closer together, and pushes instances from different classes further apart, while the third term, \clattr, encourages instances with the same protected attribute to stay apart and instances from different classes to mix together. $\alpha$ and $\beta$ are hyperparameters that control the relative importance of the cross entropy and contrastive learning terms.
	
	We experiment with two versions of the presented model. $\contrastive$ learns all components in an end-to-end fashion. 
	In addition, we present a pipelined setup, where we first train the \encoder{$\cdot$} module using the two contrastive loss components ${\clmain}-{\clattr}$, and then use its 
	output to train a logistic classifier for the main classification task. 
	This method is denoted as $\tuneboth$. 
	
	Our method differs from existing debiasing methods in that fairer representations and predictions are: (1)  achieved via contrastive learning rather than data manipulation; (2) jointly trained with the base classifier, rather than removing protected attribute information through post-processing, such as with INLP \cite{Shauli:20}; and (3) obtained without the need to train an additional network, as necessary for adversarial methods \citep{Li:18}. We show in extensive experiments that our model is superior to adversarial and post-processing methods in terms of the performance--fairness tradeoff, and more efficient to train than adversarial debiasing.

	\section{Experiments}
	\label{sec:experiments}
	We vary the architecture of \backbone{$\cdot$} across different tasks, and do not finetune it during training.\footnote{For hate speech detection and activity recognition, \backbone{$\cdot$} is first pretrained or finetuned to obtain task-specific representations, and then fixed in later stages of training.}  
	The architecture of \encoder{$\cdot$} consists of two fully-connected layers with a hidden size of 300. 
	All models are trained and evaluated on the same dataset splits, and models are selected based on their performance on the development set. 
	For fair comparisons, we finetune the learning rate, batch size, and extra hyperparameters introduced by the corresponding debiasing methods for each model on each dataset. 
	Details of the hyperparameters for each model and dataset, such as the number of layers and activation functions, are included in Supplementary Material. 
	For all experiments, we use the {A}dam optimiser \cite{Kingma:15} and early stopping with a patience of 5. 
	In the absence of a standardised method for performing model selection in fairness research (noting the complexity of model selection given the multi-objective accuracy--fairness tradeoff), we determine the best-achievable accuracy for a given model, and select the hyperparameter settings that minimise GAP while maintaining accuracy as close as possible to the best-achievable value (all based on the dev set). The development of a robust, reproducible, standardised model selection method is desperately needed in fairness research, and something that we plan to investigate in future work.
	
	
	\subsection{Baselines}
	
	We compare our method with various baselines: 
	\begin{enumerate}
		\item \vanilla: train \encoder{$\cdot$} with cross-entropy loss and no explicit bias mitigation.
		\item \inlp: train \encoder{$\cdot$} with cross-entropy loss, and apply iterative null-space projection (``INLP'': \citet{Shauli:20}) to the learned representations. 
		Specifically, a linear discriminator is iteratively trained over the protected attribute to project the representation onto the discriminator's null-space, thereby reducing protected attribute information from the representation. 
		\item \diffadv: jointly train \encoder{$\cdot$} with cross-entropy loss and an ensemble of 3 adversarial discriminators over the protected attribute, with
		an orthogonality constraint applied to each pair of sub-discriminators to encourage them to learn different aspects of the representations \cite{Han:21}. 
	\end{enumerate}

	%

	\subsection{Evaluation Metrics}
	\label{sec:metrics}
	
	To evaluate the performance of models on the main task, we adopt \textbf{Accuracy} for all four datasets.
	We measure model bias in a number of different ways, via bias in the model predictions or linear leakage over hidden or logit representations.

	\paragraph{True positive rate (TPR) gap} measures the difference in TPR between binary protected attribute $a$ and $\neg a$ for each main task class, defined as $\mathrm{GAP}^{\mathrm{TPR}}_{a,y} =|\mathrm{TPR}_{a,y}-\mathrm{TPR}_{\neg a,y}|, ~~y\in{Y}$, where $\mathrm{TPR}_{a,y}=\mathds{P}\{\hat{y}=y|y,a\}$.
	Here $\hat{y}$ and $y$ are the predicted and gold-standard main task labels; $Y$ is the set of main task labels; and $a$ and $\neg a$ represent the binary protected attribute value (such as \class{female} vs.\ \class{male}, or \aae vs.\ \sae). 
	$\mathrm{TPR}_{a,y}$ measures the percentage of correct predictions among instances with main task label $y$ and protected attribute $a$. 
	$\mathrm{GAP}^{\mathrm{TPR}}_{a,y}$ measures the absolute difference between the two different groups represented by the protected attribute, with a larger absolute value indicating larger bias.
	A difference of 0 indicates a fair model, as the prediction $\hat{y}$ is conditionally independent of protected attribute $a$. 
	Note that this formulation of the metric does not generalise to multiclass protected attributes, but in all four datasets used in this paper, all protected attributes are binary. 
	To be able to evaluate fairness where the main task label is multiclass, we follow \citet{De-Arteaga:19} and \citet{Shauli:20} in calculating the root mean square of $\mathrm{GAP}^{\mathrm{TPR}}_{a,y}$ over all classes $y\in Y$, to get a single score:
	\begin{equation}
	\begin{aligned}
	\text{\gap}&=\sqrt{\frac{1}{|Y|}\sum_{y\in{Y}}(\mathrm{GAP}^{\mathrm{TPR}}_{a,y}})^{2}
	\end{aligned}
	\end{equation}

	\paragraph{Linear leakage} measures the ability of a linear classifier to recover the protected attribute from a model's output hidden representations or logits. 
	\begin{enumerate}
		\item \hidden: based on the final hidden representation before the classifier layer. 
		\item \logits: based on the main task output $\hat{y}$ (logits).
	\end{enumerate}
    In each case, we train a linear-kernel SVM on outputs generated for the training instances, and measure leakage over the test instances. Lower values indicate a fairer model.
    
    
    \paragraph{\rank} is a single aggregate measure comprising model performance as well as the three fairness metrics (\gap and leakage at $\vec{h}$ and $\vec{\hat{y}}$). Before aggregation, we scale each metric to the unit interval by dividing the model-specific values by their respective maximum ($N(\cdot)$), so that normalized values reflect the performance of each model relative to the best result. Next we assign predictive performance and overall fairness equal weights. Between fairness measures, we weigh prediction bias equal to overall leakage, leading to: 
    $\text{\rank}{=}\frac{1}{2}{N}(\text{Accuracy})+\frac{1}{4}{N}(1{-}\text{\gap})+\frac{1}{8}{N}(1{-}\text{\hidden}){+}\frac{1}{8}{N}(1{-}\text{\logits})$.
    The best achievable \rank is 1, which indicates that a model outperformed all other models with respect to all metrics.

	\paragraph{Efficiency} measures the GPU time required to train a model to achieve the reported results, averaged over 10 runs. \\
	
	We apply our models across 4 datasets, covering NLP and vision tasks, in the form of both binary and multi-class main task classification problems. We report results in terms of accuracy, fairness (GAP and linear leakage), and efficiency across all tasks. We additionally explore the accuracy--fairness tradeoff in detail for one binary NLP task (\moji) and one multi-class computer vision task (\imsitu).

		\begin{table*}[!t]  
		\centering   	
		\scalebox{1}{ 
			\begin{threeparttable}
				\begin{tabular}{llcccccc}  
					\toprule  
					\textbf{Dataset} &\textbf{Model} &\textbf{\accuracy}$\uparrow$ &\textbf{\gap}$\downarrow$	 &\textbf{\hidden}$\downarrow$  &\textbf{\logits}$\downarrow$ &\textbf{\rank}$\uparrow$ &\textbf{\gputime}$\downarrow$\\   \midrule
					\multirow{5}{*}{\moji}
					&\vanilla &72.09$\pm$0.65  &40.21$\pm$1.23    &85.75$\pm$0.46  &70.96$\pm$2.11 &0.77 &1.0$\times$ \\				
					&\inlp &72.81$\pm$0.01  &36.81$\pm$3.49    &68.15$\pm$1.98 &67.80$\pm$1.80 &0.84 &--	\\		
					\vspace{2mm}				
					&\diffadv  &74.47$\pm$0.68 &30.59$\pm$2.94    &81.98$\pm$2.90  &65.04$\pm$1.49 &0.84  &6.5$\times$ \vspace{-1ex}\\
					&\tuneboth   &\textbf{75.99}$\pm$0.20   &14.40$\pm$1.83   &57.01$\pm$2.41 &55.42$\pm$1.14    &0.99 &0.2$\times$\\	
					&\contrastive  &75.84$\pm$0.16 &\textbf{13.92}$\pm$0.44    &\textbf{55.75}$\pm$0.21 &\textbf{55.32}$\pm$0.25 &\textbf{1.00} &1.5$\times$  	\\
					\midrule
					
					\multirow{5}{*}{\textbf{Hate speech}}
					&\vanilla &\textbf{85.04}$\pm$0.04 &18.03$\pm$0.20  &62.70$\pm$0.25 &54.51$\pm$3.10  &0.95  &1.0$\times$ \\
					&\inlp &83.76$\pm$0.12  &\textbf{13.35}$\pm$0.46     &56.81$\pm$2.61  &\textbf{52.38}$\pm$0.11 &0.98  &--	\\	
					\vspace{2mm}	
					&\diffadv  &84.00$\pm$1.39 &17.34$\pm$3.45    &58.44$\pm$1.44 &53.22$\pm$1.75 &0.96  & 7.6$\times$ \vspace{-1ex}\\
					&\tuneboth   &83.99$\pm$0.35   &14.01$\pm$0.70     &\textbf{52.83}$\pm$0.24   &52.57$\pm$0.18  &\textbf{0.99} &0.6$\times$  \\
					&\contrastive &83.56$\pm$0.48   &13.49$\pm$1.80   &\textbf{52.84}$\pm$1.80 &52.51$\pm$0.45     &\textbf{0.99}   &1.9$\times$   \\	
					\midrule
					
					\multirow{5}{*}{\bios}
					&\vanilla &\textbf{82.19}$\pm$0.04 &16.68$\pm$0.46   &99.24$\pm$0.05  &92.72$\pm$0.85 &0.76  & 1.0$\times$\\
					&\inlp  &79.42$\pm$0.28  &15.45$\pm$1.05  &92.77$\pm$6.22 &67.01$\pm$0.77  &0.85  &-- \\		
					\vspace{2mm}
					&\diffadv  &79.72$\pm$1.02 &16.78$\pm$0.87   &71.41$\pm$7.44  &69.54$\pm$6.62  &0.92  &2.8$\times$  \vspace{-1ex}\\
					&\tuneboth   &56.57$\pm$0.97  &\textbf{~7.35}$\pm$1.18   &\textbf{66.66}$\pm$2.29   &\textbf{61.06}$\pm$1.34   &0.84  &0.2$\times$   \\
					&\contrastive  &81.69$\pm$0.07  &16.83$\pm$0.36  &75.20$\pm$1.10  &66.38$\pm$1.12 &\textbf{0.93}  &0.9$\times$  	\\	
					\midrule
					
					\multirow{5}{*}{\imsitu}
					&\vanilla &\textbf{58.97}$\pm$0.66 &11.77$\pm$0.73 &72.78$\pm$0.70  &64.96$\pm$0.30 &0.94  &1.0$\times$ \\
					&\inlp &57.36$\pm$0.47  &10.53$\pm$0.87  &\textbf{60.10}$\pm$2.04  &59.06$\pm$0.38 &0.97  &--	\\	
					\vspace{2mm}	
					&\diffadv  &58.38$\pm$0.50 &10.58$\pm$0.60   &67.31$\pm$0.94 &62.37$\pm$0.73 &0.97 & 4.5$\times$ \vspace{-1ex}\\
					&\tuneboth   &57.67$\pm$0.30   &\textbf{9.41}$\pm$1.12    &71.04$\pm$0.83  &\textbf{58.34}$\pm$0.47 &0.97   &0.1$\times$ \\
					&\contrastive  &57.14$\pm$0.83  &10.41$\pm$0.77 &64.44$\pm$1.37 &59.51$\pm$1.47  &\textbf{0.98} &0.9$\times$ 				\\
					\bottomrule	
				\end{tabular}
		\end{threeparttable}}
		\caption{Experimental results on the four datasets (averaged over 10 runs). The best result for each dataset is indicated in bold. 
		Here, $\uparrow$ and $\downarrow$ indicate that higher and lower performance, resp., is better for the given metric.
		}	
		\label{results}
	\end{table*}
	
	\subsection{Experiment 1: Twitter Sentiment Analysis}
	\subsubsection{Task and Dataset}
	\label{moji_task}
	
	The task is to predict the binary sentiment for a given English tweet, based on the dataset of \citet{Blodgett:16} (\moji hereafter), where each tweet is also annotated with a binary private attribute indirectly capturing the race of the tweet author as either African American English (\aae) or Standard American English (\sae). 
	Following previous studies \cite{Shauli:20,Han:21}, the training dataset is balanced with respect to both sentiment and race but skewed in terms of sentiment--race combinations (40\% \happy-\aae, 10\% \happy-\sae, 10\% \sad-\aae, and 40\% \sad-\sae, respectively).
	\footnote{Note that the dev and test set are balanced in terms of sentiment--race combinations.} 
	The number of instances in the training, dev, and test sets are 100k, 8k, and 8k, respectively. 
	
	\subsubsection{Implementation Details}
	
	Following previous work \cite{Elazar:18,Shauli:20,Han:21}, we use \deepmoji \cite{Felbo:17}, a model pretrained over 1.2 billion English tweets, as \backbone{$\cdot$} to obtain text representations. The parameters of \deepmoji are fixed in our experiments.

	\subsubsection{Results}
	\tabref{results} (\moji) presents the results. 
	Compared to the \vanilla model, \inlp moderately reduces model bias across all metrics while retaining comparable accuracy, and \diffadv improves main task accuracy compared to \vanilla while simultaneously reducing model bias.
	Both versions of our model, \contrastive and \tuneboth, lead to the largest gain in accuracy and also the largest bias reduction across all metrics by a large margin. With leakage scores around 55, our model approaches the lower-bound value of 50 (indicating that an attacker would guess the binary protected attribute at exactly chance level).
        Overall, our methods achieve the best accuracy--fairness tradeoff.
	It is encouraging to see that incorporating debiasing techniques can contribute to improvement on the main task. 
	We hypothesise that incorporating debiasing techniques (either in the form of adversarial training or contrastive loss) acts as a form of regularisation, leading to greater robustness over the training dataset skew relative to the unbiased test set.

\begin{figure}[!t]
		\scalebox{1.0}{
			\includegraphics[width=\linewidth]{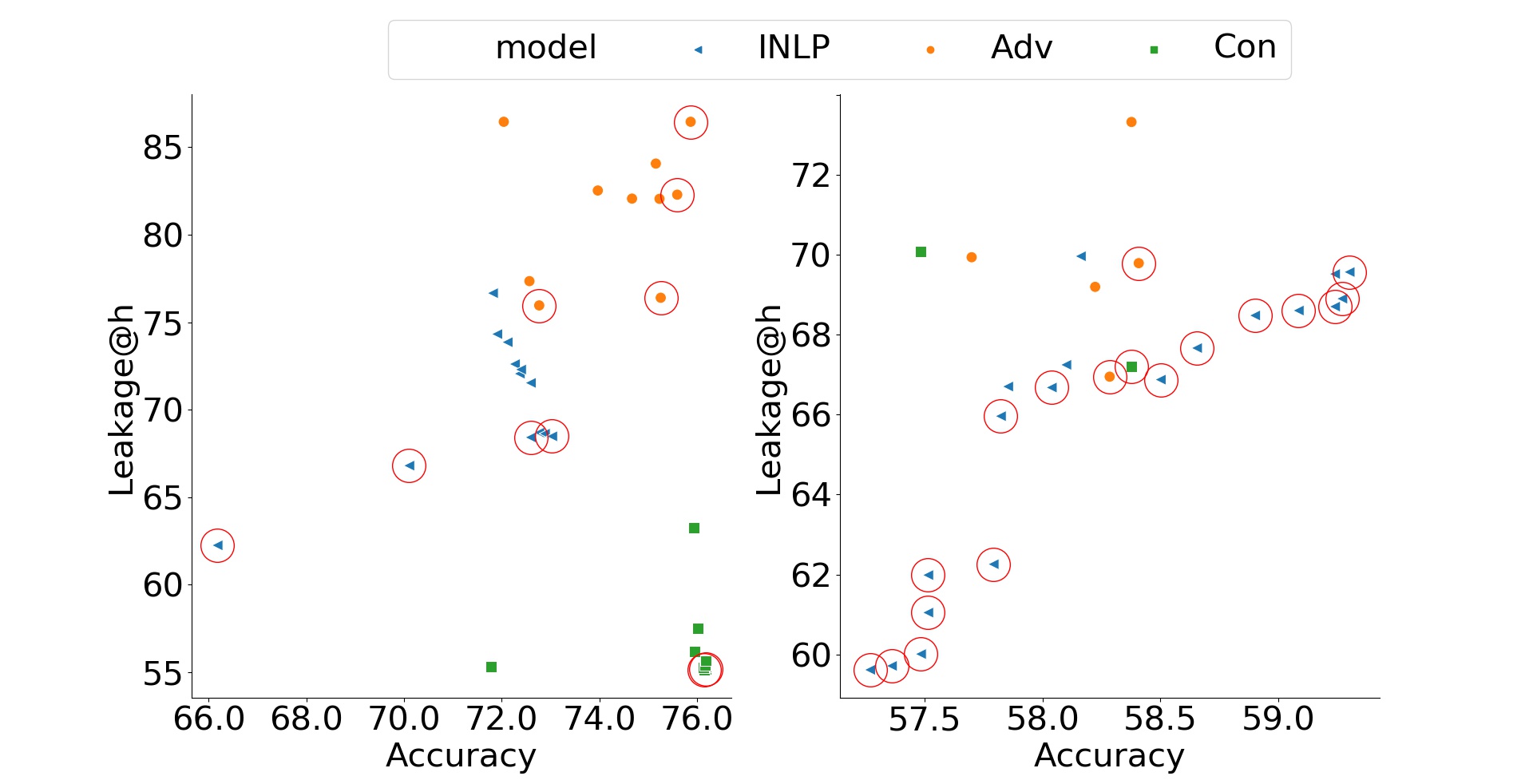}
		}
		\caption{\accuracy vs.\ \hidden of different models on the \moji (left) and \imsitu (right) test set, as we vary the most sensitive hyperparameter for each model. 
			Note that points in red circles are pareto-optimal for each model. 
		}
		\label{pareto}
	\end{figure}

	\paragraph{Accuracy--Fairness tradeoff.} We plot the tradeoff between \accuracy and \hidden for \inlp, \diffadv, and \contrastive on the test set in \figref{pareto}, where points in red circles are \pareto frontiers for each model.\footnote{Given two predictions whose \accuracy and \hidden are ($a_{1}$, $b_1$) and ($a_{2}$, $b_{2}$), if $a_{1}>a_{2}$ and $b_{1}<b_{2}$, we say the prediction ($a_{2}$, $b_{2}$) is dominated by the prediction ($a_{1}$, $b_{1}$); otherwise, they are non-dominated predictions, and form part of the \pareto frontier.} 
	The results are obtained by varying the most-sensitive hyperparameter for each model: the number of iterations for \inlp, the weight for adversarial loss for \diffadv, and $\beta$ for our method \contrastive. We can see that our proposed method achieves the best performance in terms of both \accuracy and \hidden, while \inlp and \diffadv achieve better \hidden at the cost of \accuracy. 
	

	\subsection{Experiment 2: Hate Speech Detection}
	
	\subsubsection{Task and Dataset}
	
	The task is to predict whether a tweet contains hate speech, based on the English Twitter 
	dataset of \citet{Huang:20} (\hatespeech), where each instance is labelled with several binary protected author attributes: binary gender (m/f), race (white/non-white), and age ($\le$29/$>$29).
	Following the work of \citet{Huang:20} and \citet{Han:21b}, we debias models for age, which was shown to lead to the highest model bias in prior work. 
	We use the standard data split of 31K/7K/7K train/dev/test instances, respectively. 
	
	\subsubsection{Implementation Details}
	Following \citet{Huang:20} and \citet{Han:21b}, we train a bidirectional {G}ated {R}ecurrent {U}nit \cite{Chung:14} (bi-GRU) with a cell size of 200 to predict whether a tweet contains abusive language, which is later used as \backbone{$\cdot$} to obtain text representations (400d). 
	We opt for a bi-GRU architecture as it has been shown to achieve the best bias reduction on the dataset \cite{Huang:20}.

	\subsubsection{Results}
	\tabref{results} (\hatespeech) shows the results on the test set. The vanilla \vanilla model results in less bias on this dataset compared to \moji, across all bias metrics, presumably because the train/dev/test distributions are the same for \hatespeech, but not for \moji (see Experiment 1). 
	The drop in main task performance is comparable across all methods. \inlp achieves the largest reduction in bias except for \hidden, while \diffadv decreases bias only slightly. On the other hand, \contrastive and \tuneboth achieve comparable debiasing performance with \inlp in terms of \gap, the best performance in terms of \hidden among all methods, and the best overall \rank. 
	
	
	
	
	\subsection{Experiment 3: Profession Classification}
	
	\subsubsection{Task and Dataset}
	
	The task is to predict a person's profession given their biography, based on the dataset of \citet{De-Arteaga:19}, consisting of short online biographies which have been labelled with one of 28 professions (main task label) and binary gender (protected attribute). We use the dataset split of \cite{De-Arteaga:19,Shauli:20}, consisting of 257K/40K/99K train/dev/test instances.\footnote{There are slight differences between our dataset and that used by \citet{De-Arteaga:19} and \citet{Shauli:20} as a small number of biographies were no longer available on the web when we scraped them.
	} 
	
	
	\subsubsection{Implementation Details}
	Following the work of \citet{Shauli:20}, we use the ``CLS'' token representation of the pretrained uncased \bert-base \cite{Devlin:19} as \backbone{$\cdot$}, without any further finetuning.

	\subsubsection{Results}
	\tabref{results} (\bios) shows the results on the test set. 
	We can see that \inlp achieves the best performance in terms of \gap, but the absolute bias reduction is small compared to \vanilla. 
	\contrastive achieves the best performance in terms of \logits at similar accuracy to \vanilla, while \diffadv achieves the best performance in terms of \hidden. 
	While \tuneboth substantially reduces bias across the three fairness metrics, it comes at the cost of a large drop in \accuracy, indicating the necessity of explicitly incorporating class information during training for this task. 
	Worryingly, both \diffadv and \contrastive actually marginally increase \gap.
	We hypothesise that this is because of the multi-class setting (28 classes), and the larger number of main task classes inhibiting the ability of adversarial training and contrastive learning to mitigate bias in the model under joint training.
	Overall, \contrastive once again achieves the best \rank of all the models.

	\subsection{Experiment 4: Activity Recognition}

	\subsubsection{Task and Dataset}
	
	Given an image, the task is to predict the activity depicted in the image based on the imSitu dataset \cite{Zhao:17,Wang:19}, which contains 211 activity classes and binary gender labels.
        The dataset contains roughly 110 instances for each activity, making it difficult to obtain decent performance without finetuning the backbone model, and also making debiasing impractical. 
	Therefore, we group these fine-grained labels according to their corresponding coarse-grained labels, where similar verbs are grouped into one class according to
	the {FrameNet} label hierarchy \cite{Baker+:1998}. 
	The resulting dataset contains 12 main task labels, and 12K/3K/2K train/dev/test instances.
	
	

	\subsubsection{Implementation Details}
	
	Following \citet{Wang:19}, we use standard {ResNet-50} encoder \cite{He:16} pretrained on {ImageNet}, and replace the classifier layer. 
	To extract activity-capturing representations, following the work of \citet{Wang:19}, the classifier layer is first trained with a learning rate of 0.0001 and a batch size of 128 for at most 60 epochs. 
	Then {ResNet-50} is finetuned with a learning rate of 1e-5 and a batch size of 64 for at most 60 epochs. 
	The best-performing snapshot evaluated on the dev set is used as the \backbone{$\cdot$} to obtain image representations (2,048d).

	\subsubsection{Results}
	
	\tabref{results} (\imsitu) shows the results on the test set. 
	\inlp and \diffadv decrease gap and leakage to varying degrees, with \inlp achieving better performance in terms of \logits, and \diffadv achieving better performance in terms of \hidden. 
	On the other hand, \contrastive achieves the best performance in terms of \logits and \hidden.
	Surprisingly, \tuneboth achieves the best performance in terms of \gap and \logits, which we attribute to the fact that the classifier for the main task is disconnected from the encoder training, thereby leading to better bias reduction. 
        Once again, \contrastive is best overall in terms of \rank.
	
	\paragraph{Accuracy--Fairness tradeoff.} \figref{pareto} shows the tradeoff plot between \accuracy and \hidden on the test set. 
	Different models perform differently in terms of the tradeoff, with neither \diffadv nor \contrastive reducing bias substantially over \inlp at higher levels of accuracy. 



	\subsection{Analysis}
	
	\subsubsection{Effect of Loss Components}
	
	To explore the impact of \clmain and \clattr, we conduct ablation studies on the \moji and \bios datasets by ablating one of the two contrastive loss components. 
	We denote the model trained with $\alpha\clce+\beta\text{\clmain}$ as \ceadd, and the model trained with $\alpha\clce-\beta\text{\clattr}$ as \ceminus. 
	
\begin{figure}[!t]
		\includegraphics[width=\linewidth]{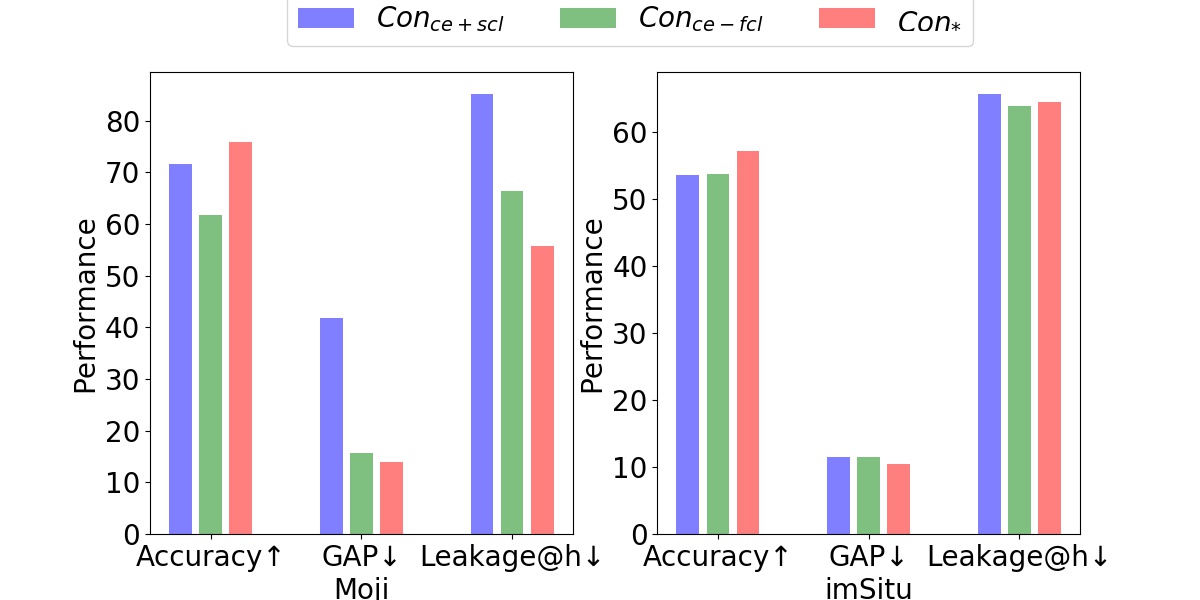}
		\caption{Effects of contrastive loss components.}
		\label{loss_component}
	\end{figure}

	The results are shown in \figref{loss_component}. 
	We can see that \contrastive achieves the best performance across all evaluation metrics on the \moji dataset. 
	On the \imsitu dataset, \contrastive also achieves the best accuracy, while roughly equalling the best bias results.
	This illustrates the advantage of incorporating both contrastive loss components. 
	
	\subsubsection{Visualising Representations}


	\begin{figure}[!t]
		\centering
		\begin{subfigure}[t!]{1\columnwidth}
			\begin{center}
				\includegraphics[width=\linewidth]{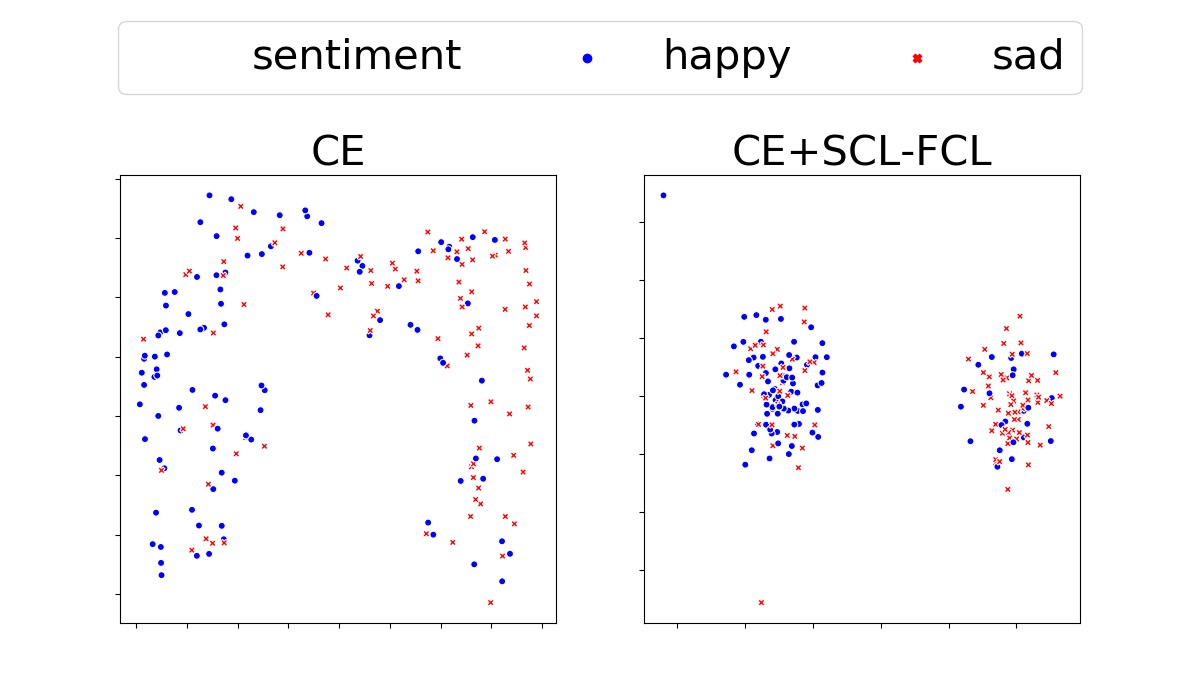}
			\end{center}
			\vspace{-3ex}
			\label{moji_sentiment}
		\end{subfigure}
		\begin{subfigure}[t!]{1\columnwidth}
			\begin{center}
				\includegraphics[width=\textwidth]{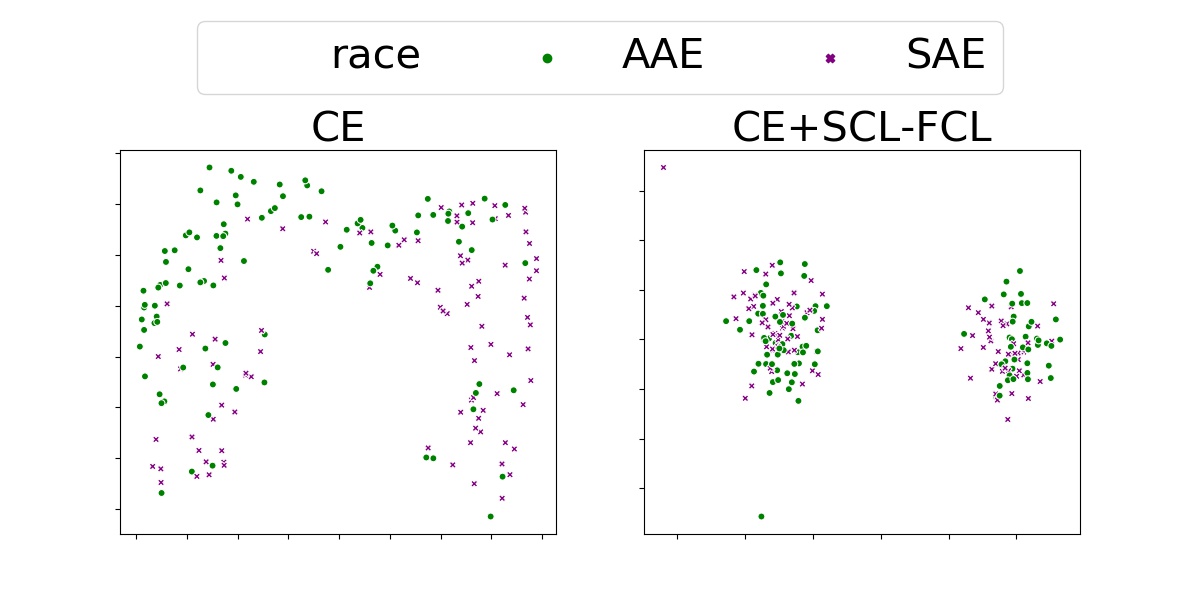}
			\end{center}
			\vspace{-3ex}
			\label{moji_race}
		\end{subfigure}
 			\caption{t-SNE scatter plots of learned representations of \vanilla and \contrastive over the Moji dataset (based on 100 random samples from each main task class; best viewed in colour). Red and blue colours indicate that they have different sentiment (main task) labels: red $\rightarrow$ \sad and blue $\rightarrow$ \happy. Green and purple colours indicate that they have different race groups (protected attribute): green $\rightarrow$ \class{\sae} and purple $\rightarrow$ \class{\aae}.
		}
		\label{moji_tsne}
	\end{figure}
	
	
	In \figref{moji_tsne}, we show t-SNE plots of the learned representations of \vanilla and \contrastive on the \moji training set from the perspectives of the main task labels and protected attribute values. 
	We can clearly see that for \vanilla, the positive (\happy) instances are mostly on the left of the figure and negative (\sad) instances are mostly on the right of \figref{moji_tsne} (upper left figure). 
	From the race perspective, \aae instances are more likely towards the left and instances with \sae are most likely to be towards the right of \figref{moji_tsne} (bottom left figure). 
	For \contrastive, the resulting representations show that instances belonging to the same class cluster together in terms of sentiment, and instances belonging to the different classes mix together in terms of race, affirming our motivation. 
	
	\subsubsection{Binary Classification vs.\ Multi-class Classification}
	
	Based on experimental results from \tabref{results}, we observed that \contrastive is effective in reducing bias in terms of \gap, \logits, and \hidden on the \moji and \hatespeech datasets, which are both binary classification tasks. 
	The relatively less impressive results of \contrastive on \bios and \imsitu can be explained by the fact that it is difficult to find the sweet spot for representing instances in multi-class classification settings. 
	That is, the more classes there are, the harder it is for the optimiser to separate the classes from one another, leading to scattered representations at the main task class level, and making  contrastive learning less effective in bias reduction. 
	
	To verify this hypothesis, we focus on the professions of \nurse and \surgeon in the Bios dataset, two professions with well-documented gender stereotypes. 
	As shown in \tabref{bios_binary_results}, the row of \textsf{CE}$^{m}$ presents the results of the instances belonging to the class ``nurse'' and ``surgeon'' in the multi-class classification setting and the row of \textsf{CE}$^{b}$ shows the results of the same set data in the binary classification setting. 
	We can see that moving from the multi-class classification setting into the binary classification setting, there is a big increase in \accuracy and a small decrease in \gap and \logits. 
	In the binary classification setting, neither \inlp nor \diffadv can effectively reduce \gap, while \contrastive is much more effective at reducing gender bias across all bias evaluation metrics in the binary classification setting, with a small decrease in \accuracy. 

	\begin{table}[!t]  
	\centering   	
	\scalebox{0.8}{ 
		\begin{threeparttable}
			\begin{tabular}{llccc}  
				\toprule  
				\textbf{Model} &\textbf{\accuracy}$\uparrow$ &\textbf{\gap}$\downarrow$ &\textbf{\hidden}$\downarrow$ &\textbf{\logits}$\downarrow$  \\ \midrule
				\textsf{CE}$^{m}$  &73.64$\pm$0.98   &21.60$\pm$0.75  &98.98$\pm$0.05  &93.67$\pm$0.38\\ \midrule
				\textsf{CE}$^{b}$   &\textbf{96.60}$\pm$0.04   &16.72$\pm$0.24  &99.06$\pm$0.04  &90.29$\pm$0.76\\
				\inlp  &96.58$\pm$0.09  &15.84$\pm$0.78   & 92.51$\pm$1.71  & 89.40$\pm$0.08 \\
				\diffadv  &96.38$\pm$0.39  &10.41$\pm$2.04  &89.99$\pm$2.04 &87.85$\pm$0.41  \\ 	
				\contrastive  &94.56$\pm$0.16 &\textbf{4.88}$\pm$0.46  &\textbf{85.61}$\pm$0.32 &\textbf{85.61}$\pm$0.32 	\\
				
				\bottomrule		
				\end{tabular}
				\end{threeparttable}}
			\caption{Results of \contrastive in the binary classification setting (nurse vs. surgeon) on the \bios test set.}
			\label{bios_binary_results}
	\end{table} 
	

	\subsection{Limitations}
	
	A limitation of our proposed approach is that the method is designed to remove information related to protected attributes based on the assumption that the attacker model will be a linear classifier. 
	We leave the investigation of protecting against attacks by non-linear classifiers to future work. 
	In our work, the \backbone{$\cdot$} is not learned or fine-tuned together with \encoder{$\cdot$} and the classification layer in an end-to-end fashion. 
	However, finetuning the \backbone{$\cdot$} has the potential for better task-specific or semantic-preserving representations of text and images, which may further remove biases encoded in the the pretrained models. 
	Exploring how to construct informative negative samples in the multi-class classification setting, which has not been studied in the literature, is also an interesting direction for future work.

	\section{Conclusion}
	\label{sec:conclusion}
	
	Biased representations and predictions can reinforce existing societal biases and stereotypes. 
	Based on the intuition that similar instances belonging to the same main task class should be pulled together and similar instances belonging to the same protected attribute class should be pushed apart in the representation space, we proposed to combine cross-entropy loss with two contrastive loss components in optimising neural networks. 
	Experimental results over four NLP and vision datasets demonstrate the effectiveness of our proposed method. 
	Further analysis and ablation studies indicate the necessity of incorporating both contrastive loss components in bias reduction, to maintain main task accuracy.

	\bibliography{aaai22}	
\end{document}